\documentclass[conference]{IEEEtran}
\IEEEoverridecommandlockouts
\usepackage{cite}
\usepackage{placeins}
\usepackage{amsmath,amssymb,amsfonts}
\usepackage{algorithmic}
\usepackage{graphicx}
\usepackage{textcomp}
\usepackage{xcolor}
\usepackage{graphicx}
\usepackage{booktabs,multirow,tabularx}
\usepackage{adjustbox}
\usepackage{float}
\usepackage{wrapfig}
\usepackage[rightcaption,raggedright]{sidecap}
\usepackage{colortbl}
\usepackage[ruled,vlined,linesnumbered,longend]{algorithm2e}
\usepackage{graphicx}
\usepackage{subcaption}
\usepackage{placeins}
\SetKwInput{KwIn}{Input}
\SetKwInput{KwOut}{Output}
\SetNlSty{}{}{:}
\SetAlgoNlRelativeSize{0}
\SetInd{0.5em}{1em}

\usepackage[hidelinks]{hyperref}
\usepackage{cleveref}
\def\BibTeX{{\rm B\kern-.05em{\sc i\kern-.025em b}\kern-.08em
    T\kern-.1667em\lower.7ex\hbox{E}\kern-.125emX}}
\begin{document}

\title{Robust Incomplete Multimodal Sentiment Analysis via Iterative Proxy Correction\\
\thanks{\IEEEauthorrefmark{2}Subin Huang and Chao Kong are with Bigdata and Responsible Artificial Intelligence for National Governance, Renmin University of China, Beijing 100872, China.}
\thanks{This work was also supported in part by the Anhui Provincial Natural Science Foundation under Grant No. 2308085MF220.}
}

\author{

\IEEEauthorblockN{1\textsuperscript{st} Zhifa Geng}
\IEEEauthorblockA{
School of Computer and Information\\
Anhui Polytechnic University\\
Wuhu, China\\
zhifageng77@gmail.com\\
}
\and

\IEEEauthorblockN{2\textsuperscript{nd} Subin Huang\IEEEauthorrefmark{1}\IEEEauthorrefmark{2}}  \thanks{* Subin Huang is the corresponding author.}

\IEEEauthorblockA{
School of Computer and Information\\
Anhui Polytechnic University\\
Wuhu, China\\
subinhuang@ahpu.edu.cn\\
}
\and

\IEEEauthorblockN{3\textsuperscript{rd} Hao Guo}
\IEEEauthorblockA{
School of Computer and Information\\
Anhui Polytechnic University\\
Wuhu, China\\
sifan10077@gmail.com\\
}
\and

\IEEEauthorblockN{4\textsuperscript{th} Junjie Chen}
\IEEEauthorblockA{
School of Computer and Information\\
Anhui Polytechnic University\\
Wuhu, China\\
jorji.chen@gmail.com\\
}
\and

\IEEEauthorblockN{5\textsuperscript{th} Sanmin Liu}
\IEEEauthorblockA{
School of Computer and Information\\
Anhui Polytechnic University\\
Wuhu, China\\
sanmin.liu@ahpu.edu.cn\\
}
\and

\IEEEauthorblockN{6\textsuperscript{th} Chao Kong\IEEEauthorrefmark{2}}
\IEEEauthorblockA{
School of Computer and Information\\
Anhui Polytechnic University\\
Wuhu, China\\
kongchao@ahpu.edu.cn\\
}

}

\maketitle
\begin{abstract}

Multimodal sentiment analysis aims to infer affective states by integrating language, visual, and acoustic cues. However, real-world multimodal inputs are often incomplete or corrupted, which can weaken cross-modal complementarity and introduce misleading information into downstream fusion. Existing proxy-based methods for incomplete MSA commonly rely on one-shot proxy construction to compensate for degraded language information, but the generated proxy may be coarse or unreliable at initialization. Prematurely injecting such a proxy into multimodal reasoning can propagate initial errors and compromise sentiment prediction. To address this limitation, we propose an iterative proxy correction framework for robust incomplete MSA. Our method constructs a language-oriented proxy from non-language modalities and progressively refines it under multimodal context through gated residual correction. The corrected proxy is then adaptively fused with the observed language representation according to an estimated language reliability score, allowing the model to balance proxy-based compensation and trustworthy linguistic evidence. In addition, we introduce a stage-wise latent correction objective that uses the complete language representation as a training-time semantic anchor to stabilize the proxy refinement trajectory. Extensive experiments on MOSI, MOSEI, and SIMS under diverse missing-modality settings demonstrate that the proposed framework consistently outperforms competitive baselines and achieves robust sentiment prediction under incomplete inputs.

\end{abstract}


\section{Introduction}
Multimodal sentiment analysis (MSA) aims to infer human affective states by integrating complementary cues from language, vision, and acoustics~\cite{li2025tactic}. Recent MSA models exploit cross-modal interactions to achieve strong performance, but usually assume that all modalities are complete and reliable~\cite{liu2026,AFROZE2026,Wang-2025-PublicBehavior}. In real-world scenarios, however, multimodal observations are often incomplete or corrupted due to background noise, sensor failures, transmission instability, or privacy constraints. Under such conditions, the challenge goes beyond simple information loss: degraded modalities may still participate in cross-modal interaction and inject misleading information into multimodal fusion, thereby distorting multimodal representations and biasing sentiment prediction.


To improve robustness under incomplete inputs, existing methods for incomplete MSA mainly focus on compensating for missing or degraded modality information, and can be broadly divided into reconstruction-based and representation-based paradigms. Reconstruction-based methods~\cite{ZHAI2026131746,ZHAO2025126974,zheng20263d,guo2024multimodal} explicitly recover missing signals or features before prediction, while representation-based methods~\cite{GONG2026111870,lin-etal-2025-t2dr,wang2025sslmm} directly learn decision-oriented latent representations without raw-signal recovery. Within the latter paradigm, language-centered methods are widely adopted because textual information usually provides explicit and task-relevant semantic cues~\cite{li2024unified,hazarika2020misa}. However, this strategy implicitly assumes that language representations are sufficiently reliable, which may not hold when textual inputs are missing, noisy, or semantically insufficient. This has motivated language-oriented proxy-based methods~\cite{zhu-etal-2025-proxy,zhang2024towards}, which construct auxiliary latent representations to compensate for degraded language information while preserving language-guided sentiment inference.

\begin{figure}[t]
    \centering
    \includegraphics[width=0.98\linewidth]{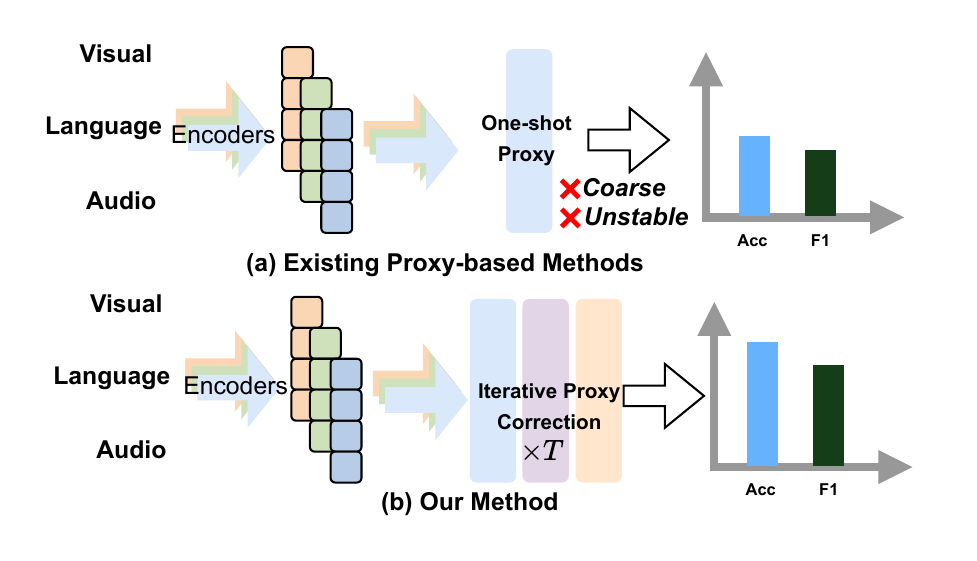}
    \caption{Comparison between existing proxy-based methods and our iterative proxy correction method.}
    \label{fig1}
\end{figure}

However, as illustrated in Figure~\ref{fig1}(a), many existing proxy-based methods follow a one-shot construction paradigm, where the proxy is directly injected into subsequent multimodal interaction once it is generated. This paradigm implicitly assumes that the initialized proxy is sufficiently reliable. However, under incomplete conditions, the proxy is often inferred from degraded observations and may remain coarse, noisy, or only partially informative. Prematurely involving such an imperfect proxy in cross-modal fusion can propagate and amplify its initial errors during subsequent reasoning, ultimately degrading sentiment prediction.

To address this issue, we propose an iterative proxy correction framework for incomplete multimodal sentiment analysis, as shown in Figure~\ref{fig1}(b). Instead of treating the proxy as a fixed substitute, we regard it as a language-oriented auxiliary representation that should be progressively refined before downstream decision-making. Specifically, the proxy is first initialized from non-language modalities and then iteratively updated under multimodal context through gated residual correction. The corrected proxy is further adaptively fused with the observed language representation according to its estimated reliability before downstream reasoning. Moreover, we introduce a stage-wise latent correction objective that uses the complete language representation only during training as a semantic anchor, encouraging the proxy to become progressively more stable and informative throughout the refinement process. The main contributions of this work are summarized as follows:

\begin{itemize}
    \item We identify the error-propagation risk of one-shot proxy construction in incomplete MSA and propose iterative proxy correction to improve proxy reliability before downstream fusion.
    
    \item We develop a gated residual correction mechanism to progressively refine a language-oriented proxy under multimodal context, guided by a stage-wise latent correction objective.
    
    \item Extensive experiments under diverse incomplete-modality settings demonstrate the robustness and effectiveness of our method against competitive incomplete MSA baselines.
\end{itemize}

\section{Related Work}



In real-world scenarios, multimodal inputs are often incomplete due to sensor failure, noise, occlusion, or privacy constraints. Existing studies on incomplete multimodal learning mainly fall into two categories:  reconstruction-based methods and representation-based methods. 

\textbf{Reconstruction-based methods.} The first line reconstructs absent signals through cross-modal generation or imputation. Early efforts explored multimodal autoencoding and adversarial generation for missing-view completion~\cite{Luan2017cascaded,shang2017vigan}. More recent MSA methods exploit semantic correlations across modalities for targeted recovery. For example, MPLMM~\cite{guo2024multimodal} adopts prompt-based generation for lightweight recovery, TRML~\cite{ZHAO2025126974} generates semantically aligned virtual missing modalities based on CLIP, and HyPLe-MKD~\cite{ZHAI2026131746} combines hybrid prompt learning with multilevel distillation for missing-modality generation and fusion.

\textbf{Representation-based methods.} The second line avoids explicit reconstruction and instead emphasizes robust fusion under incomplete inputs. Representative methods improve incomplete-modality learning through adaptive fusion and self-distillation~\cite{zeng2022robust,li2024unified}. Recent studies further enhance this direction with proxy modality, prototype guidance, and self-supervised pre-training~\cite{zhu-etal-2025-proxy,wang2025sslmm}. For instance, MIG-HCL~\cite{GONG2026111870} builds uni- and multimodal interaction graphs with hybrid contrastive learning, while T$^2$DR~\cite{lin-etal-2025-t2dr} improves incomplete multimodal learning through deficiency-resistant attention, shared feature prediction, and capability-aware fusion under mixed missing scenarios.

\section{Method}

\subsection{Multimodal Input}
\begin{figure*}
    \centering
    \includegraphics[width=0.971\linewidth]{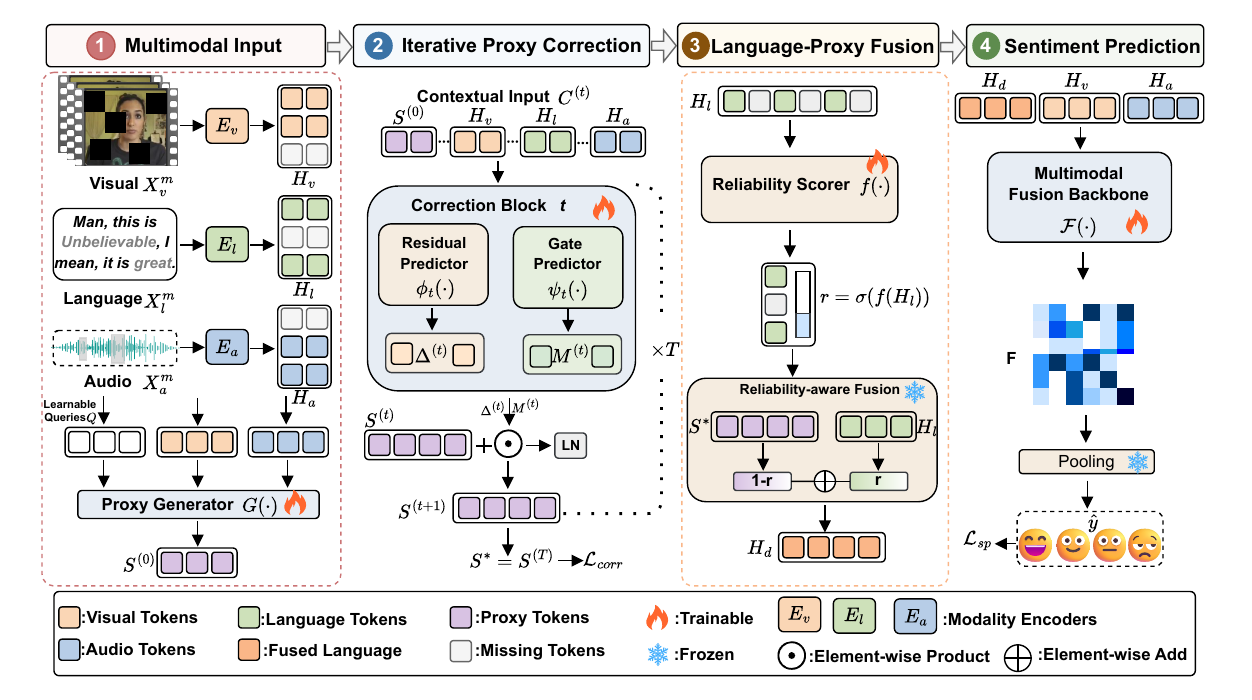}
    \caption{Overall architecture of the proposed method. }
    \label{fig2}
\end{figure*}

The overall pipeline for the proposed iterative proxy correction framework for incomplete MSA is illustrated in Figure~\ref{fig2}. Given an incomplete multimodal utterance $\mathcal{X}^{m}=\{X_l^{m},X_v^{m},X_a^{m}\}$, the goal is to predict its sentiment label $y$. During training, the corresponding complete sample $\mathcal{X}^{c}=\{X_l^{c},X_v^{c},X_a^{c}\}$ is available only for auxiliary supervision. We encode the incomplete language, visual, and acoustic modalities into a shared latent space:
\begin{equation}
H_l^m = E_l(X_l^{m}), \quad H_v^m = E_v(X_v^{m}), \quad H_a^m = E_a(X_a^{m}),
\end{equation}
where $E_l(\cdot)$, $E_v(\cdot)$, and $E_a(\cdot)$ are modality-specific encoders. After token compression, $H_l^m,H_v^m,H_a^m \in \mathbb{R}^{N \times d}$, where $N$ is the number of tokens and $d$ is the hidden dimension.

\subsection{Iterative Proxy Correction}

Directly relying on the observed language representation $H_l^m$ can be suboptimal when important semantics are corrupted or missing. We therefore construct a language proxy in latent space. To avoid trivial re-parameterization of the observed language stream, the proxy is initialized solely from non-language modalities, encouraging it to capture complementary cues unavailable from the corrupted language input.

Let $Q \in \mathbb{R}^{N \times d}$ denote a set of learnable query tokens. The initial proxy is generated as
\begin{equation}
S^{(0)} = G([Q;H_a^m;H_v^m]),
\end{equation}
where $[\cdot;\cdot]$ denotes token-wise concatenation and $G(\cdot)$ is implemented as a Transformer-based proxy generator. Since $H_a^m$ and $H_v^m$ are encoded from incomplete inputs, $S^{(0)}$ is only a coarse estimate from available non-language evidence.

To improve this estimate, we refine the proxy iteratively under multimodal context. At the $t$-th step, we form
\begin{equation}
C^{(t)} = [S^{(t)};H_l^m;H_a^m;H_v^m].
\end{equation}
A residual correction term and an update gate are then computed as
\begin{equation}
\Delta^{(t)} = \phi_t(C^{(t)}), \qquad
M^{(t)} = \sigma(\psi_t(C^{(t)})),
\end{equation}
where $\phi_t(\cdot)$ and $\psi_t(\cdot)$ are learnable mappings, $\sigma(\cdot)$ is the sigmoid function, and both $\Delta^{(t)}$ and $M^{(t)}$ have the same shape as $S^{(t)}$. The proxy is updated by
\begin{equation}
S^{(t+1)} = \mathrm{LN}\!\left(S^{(t)} + M^{(t)} \odot \Delta^{(t)}\right),
\end{equation}
where $\odot$ denotes element-wise multiplication and $\mathrm{LN}(\cdot)$ denotes layer normalization. After $T$ correction steps, the final corrected proxy is
\begin{equation}
S^{*}=S^{(T)}.
\end{equation}


\subsection{Language-Proxy Fusion}

Although the corrected proxy $S^{*}$ provides complementary semantic evidence, the observed language representation $H_l^m$ may still retain useful sentiment cues. We therefore adaptively fuse them using an estimated language reliability score:
\begin{equation}
r = \sigma(f(\mathrm{Pool}(H_l^m))),
\end{equation}
where $f(\cdot)$ is a scoring network and $r \in [0,1]$ measures the estimated reliability of the observed language representation. The final fused language representation is computed as
\begin{equation}
H_d = (1-r)S^{*} + rH_l^m.
\end{equation}
In this way, the model trusts $H_l^m$ more when it is reliable, and otherwise shifts more weight to the corrected proxy.

\subsection{Sentiment Prediction}

The fused language representation is integrated with the visual and acoustic features through the downstream multimodal reasoning backbone:
\begin{equation}
F = \mathcal{F}(H_d,H_v^m,H_a^m),
\end{equation}
where $\mathcal{F}(\cdot)$ denotes the inherited fusion network. The final sentiment prediction is obtained as
\begin{equation}
\hat{y}=W_o\,\mathrm{Pool}(F)+b_o.
\end{equation}

\subsection{Training Objective}

The model is optimized with a sentiment prediction loss and a stage-wise correction loss. The prediction loss is
\begin{equation}
\mathcal{L}_{\mathrm{sp}}=\|\hat{y}-y\|_2^2.
\end{equation}

To supervise progressive proxy correction, we use the complete language representation $H_l^{c}$ as a semantic anchor:
\begin{equation}
\mathcal{L}_{\mathrm{corr}}=\sum_{t=0}^{T}\lambda_t \|S^{(t)}-H_l^{c}\|_2^2,
\end{equation}
where $\sum_{t=0}^{T}\lambda_t=1$ and larger weights are assigned to later stages.

Following the base framework, we keep the remaining auxiliary objectives unchanged and summarize them as $\mathcal{L}_{\mathrm{aux}}$, with implementation details following~\cite{zhang2024towards}. The final loss is
\begin{equation}
\mathcal{L}=\mathcal{L}_{\mathrm{sp}}+\theta \mathcal{L}_{\mathrm{corr}}+\mathcal{L}_{\mathrm{aux}},
\end{equation}
where $\theta$ controls the strength of the correction supervision.

\section{Experiment}
\begin{table*}[t]
\centering
\caption{Results on the MOSI and MOSEI datasets (baseline results from LNLN~\cite{zhang2024towards}). }
\label{tab:results1}
\resizebox{\textwidth}{!}{%
\begin{tabular}{lcccccc cccccc}
\toprule
\multirow{2}{*}{Method} & \multicolumn{6}{c}{\textbf{MOSI}} & \multicolumn{6}{c}{\textbf{MOSEI}} \\
\cmidrule(lr){2-7}\cmidrule(lr){8-13}
& Acc-7  & Acc-5  & Acc-2  & F1  & MAE & Corr 
& Acc-7 & Acc-5  & Acc-2  & F1 & MAE  & Corr \\
\midrule
MISA    & 29.85 & 33.08 & 71.49/70.00 & 71.28/70.33 & 1.085 & 0.524 & 40.84 & 39.39 & 71.27/75.82 & 63.85/68.73 & 0.780 & 0.503 \\
Self-MM & 29.55 & 34.67 & 70.51/69.26 & 66.60/67.54 & \underline{1.070} & 0.512 & 44.70 & 45.38 & 73.89/77.42 & 68.92/72.31 & 0.695 & 0.498 \\
MMIM    & 31.30 & 33.77 & 69.14/67.06 & 66.65/64.04 & 1.077 & 0.507 & 40.75 & 41.74 & 73.32/75.89 & 68.72/70.32 & 0.739 & 0.489 \\
CENET   & 30.38 & 37.25 & 71.46/67.73 & 68.41/64.85 & 1.080 & 0.504 & \textbf{47.18} & \underline{47.83} & 74.67/77.34 & 70.68/74.08 & 0.685 & \underline{0.535} \\
TETFN   & 30.30 & 34.34 & 69.76/67.68 & 65.69/63.29 & 1.087 & 0.507 & 40.30 & 47.70 & 69.76/67.68 & 65.69/63.29 & 1.087 & 0.508 \\ 
ALMT    & 30.30 & 33.42 & 70.40/68.39 & 72.57/\underline{71.80} & 1.083 & 0.498 & 40.92 & 41.64 & \underline{76.64}/77.54 & 77.14/78.03 & \underline{0.674} & 0.481 \\
LNLN    & \underline{34.26} & \underline{38.27} & \underline{72.55}/\underline{70.94} & \underline{72.73}/71.25 & \textbf{1.046} & \underline{0.527} & 45.42 & 46.17 & 76.30/\underline{78.19} & \underline{77.77}/\underline{79.95} & 0.692 & 0.530 \\
\midrule
\rowcolor{green!10}
\textbf{Ours} & \textbf{34.52} & \textbf{38.55} & \textbf{73.27}/\textbf{71.93} & \textbf{73.03}/\textbf{71.93} & \textbf{1.046} & \textbf{0.532} 
& \underline{47.10} & \textbf{47.94} & \textbf{78.20}/\textbf{78.94} & \textbf{78.46}/\textbf{80.00} & \textbf{0.664} & \textbf{0.595} \\
\bottomrule
\end{tabular}
}
\end{table*}

\subsection{Benchmarks}
We evaluate our method on three widely used multimodal sentiment benchmarks: MOSI, MOSEI, and SIMS, covering both English and Chinese scenarios. MOSI contains 2,199 English utterances from online movie reviews, each annotated with text, audio, vision, and a sentiment score in $[-3, +3]$. MOSEI is a larger English benchmark with 22,856 utterances from more diverse speakers and topics, making it more challenging for robustness evaluation. SIMS is a Chinese benchmark with 2,281 manually annotated segments collected from movie and TV dialogues, featuring richer conversational context and more subtle sentiment expressions.

\subsection{Evaluation Metrics and Implementation Details}

We use MAE and Corr for regression evaluation. For MOSI and MOSEI, we additionally report binary classification results (Acc-2 and F1) under both negative/positive and negative/non-negative settings; when presented as ``a/b'', the two values correspond to these settings, respectively. We also report Acc-5 and Acc-7 on MOSI/MOSEI, and Acc-3 and Acc-5 on SIMS.

All models are implemented in PyTorch and optimized with Adam. The learning rate and weight decay are both set to $1\times10^{-4}$, the batch size is 64, and the default number of training epochs is 200. Experiments are conducted on a single NVIDIA RTX 4090 GPU with 8 workers. To evaluate robustness under incomplete multimodal inputs, we vary the missing rate from 0.0 to 0.9. Each setting is run three times with seeds 1111, 1112, and 1113, and we report the average results.

\subsection{Baselines} 


We compare our method with several representative multimodal sentiment models, including Self-MM~\cite{yu2021learning}, CENet~\cite{wang2022cross}, TETFN~\cite{wang2023tetfn}, MMIM~\cite{han2021improving}, MISA~\cite{hazarika2020misa}, ALMT~\cite{zhang2023learning}, and LNLN~\cite{zhang2024towards}. These baselines cover diverse design paradigms, such as unimodal-multimodal joint learning, text-centered interaction, mutual-information maximization, shared-specific disentanglement, adaptive multimodal fusion, and robustness modeling under missing modalities.

\subsection{Robustness Comparison}

Tables~\ref{tab:results1} and ~\ref{tab:results2} report the results on MOSI, MOSEI, and SIMS, where the best and second-best results are highlighted in bold and underlined, respectively. Overall, our method achieves the best or highly competitive performance across all three benchmarks, demonstrating strong effectiveness and robustness. On MOSI, our method achieves the best or tied-best results on all metrics. Compared with the strongest baseline LNLN, it consistently improves Acc-7, Acc-5, Acc-2, F1, and Corr, while matching the best MAE, showing clear advantages in both classification and regression. On MOSEI, our method obtains the best results on five out of six metrics and ranks second on Acc-7 with only a marginal gap. In particular, it achieves the best MAE and Corr, with Corr improving from 0.535 to 0.595, indicating a stronger ability to model fine-grained sentiment intensity. On SIMS, our method also delivers robust performance, achieving the best results on Acc-5, Acc-3, MAE, and Corr, while remaining competitive on Acc-2 and F1. Although it does not obtain the best F1 score, the overall results still suggest a better balance between classification accuracy and regression quality.

\begin{figure}[t]
    \centering
    \setlength{\abovecaptionskip}{2pt}
    \setlength{\belowcaptionskip}{-8pt}
    \includegraphics[width=1.0\linewidth]{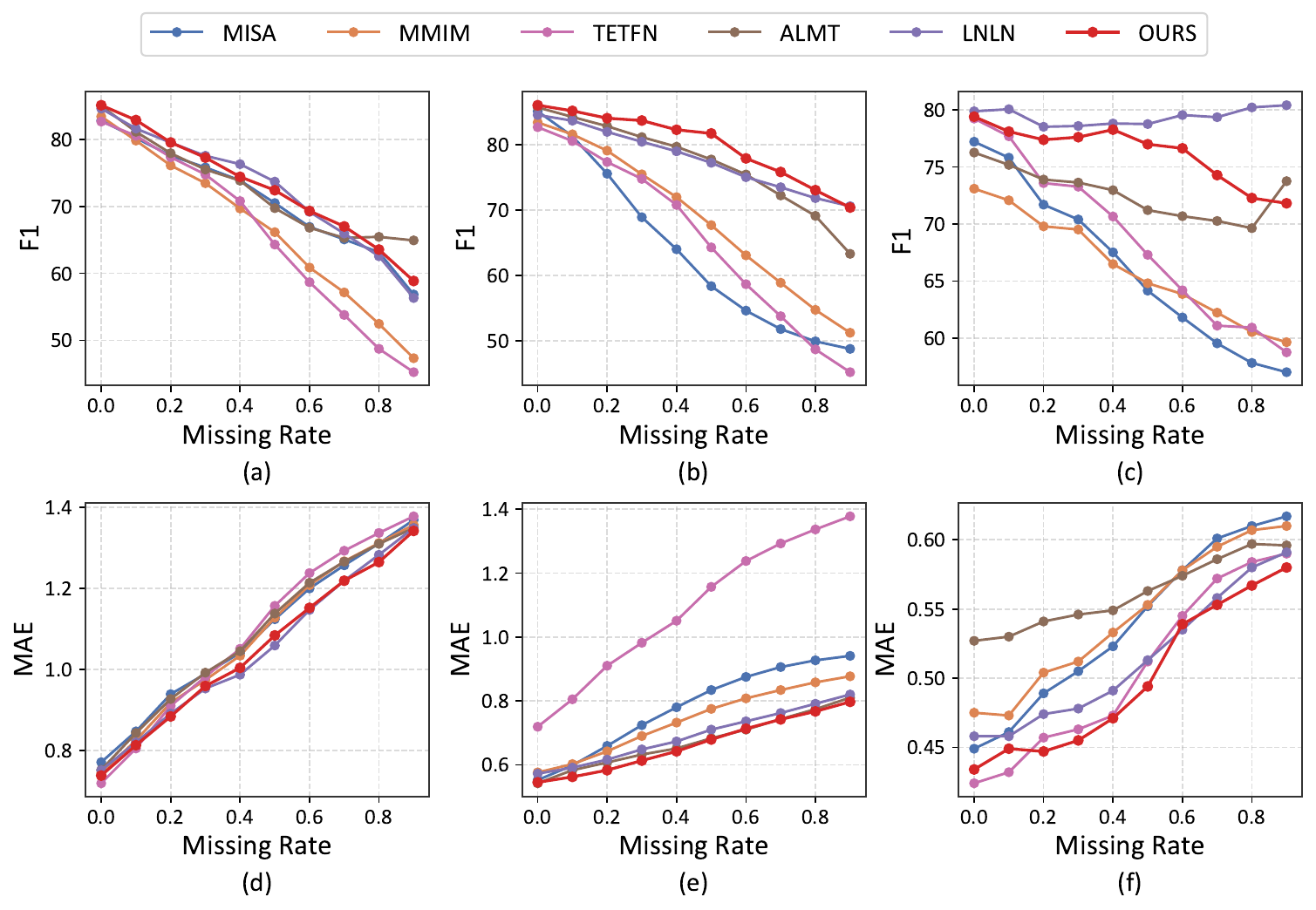}
    \caption{(a)–(c) show the F1 curves on MOSI, MOSEI, and SIMS. (d)–(f) show the MAE curves on MOSI, MOSEI, and SIMS.}
    \label{fig3}
    
\end{figure}

Figure~\ref{fig3} further shows that increasing missing rates consistently deteriorate the performance of all compared methods, highlighting the difficulty of incomplete multimodal sentiment analysis. Nevertheless, our method exhibits better resistance to such degradation and maintains more stable performance across a wide range of missing conditions.

\begin{table}[t]
\centering
\caption{Results on the SIMS dataset (baseline results from LNLN~\cite{zhang2024towards}).}
\label{tab:results2}
\setlength{\tabcolsep}{6pt}
\renewcommand{\arraystretch}{1.08}
\begin{tabularx}{\linewidth}{l*{6}{>{\centering\arraybackslash}X}}
\toprule
Method & Acc-5  & Acc-3  & Acc-2  & F1 & MAE & Corr  \\
\midrule
MISA    & 31.53 & 56.87 & 72.71 & 66.30 & 0.539 & 0.348 \\
Self-MM & 32.28 & 56.75 & 72.81 & 68.43 & 0.508 & 0.376 \\
MMIM    & 31.81 & 52.76 & 69.86 & 66.21 & 0.544 & 0.339 \\
CENET   & 22.29 & 53.17 & 68.13 & 57.90 & 0.589 & 0.107 \\
TETFN   & 33.42 & 56.91 & \textbf{73.58} & 68.67 & \underline{0.505} & 0.387 \\
ALMT    & 20.00 & 45.36 & 69.66 & 72.76 & 0.561 & 0.364 \\
LNLN    & \underline{34.64} & \underline{57.14} & 72.73 & \textbf{79.43} & 0.514 & \underline{0.397} \\
\midrule
\rowcolor{green!10}
\textbf{Ours} & \textbf{35.13} & \textbf{58.28} & \underline{73.05} & \underline{76.29} & \textbf{0.498} & \textbf{0.404} \\
\bottomrule
\end{tabularx}
\end{table}

\begin{table*}[t]
\centering
\caption{Modality ablation results on the MOSI and MOSEI datasets.}
\label{tab:results3}
\resizebox{\textwidth}{!}{%
\begin{tabular}{lcccccc cccccc}
\toprule
\multirow{2}{*}{Method} & \multicolumn{6}{c}{MOSI} & \multicolumn{6}{c}{MOSEI} \\
\cmidrule(lr){2-7}\cmidrule(lr){8-13}
& Acc-7  & Acc-5  & Acc-2  & F1  & MAE & Corr 
& Acc-7 & Acc-5  & Acc-2  & F1 & MAE  & Corr \\
\midrule
T   & 45.58 & 51.70 & 84.91\,/\,82.75 & 84.84\,/\,82.68 & 0.731 & 0.790 & 52.39 & 53.81 & 85.75\,/\,84.46 & 85.7\,/\,84.14 & 0.548 & 0.769 \\
A   & 22.84 & 23.08 & 58.49\,/\,57.38 & 61.27\,/\,59.64 & 1.371 & 0.280 & 41.38 & 41.38 & 63.84\,/\,71.28 & 51.85\,/\,59.93 & 0.830 & 0.152 \\
V   & 22.98 & 24.83 & 58.59\,/\,56.80 & 51.78\,/\,53.90 & 1.376 & 0.230 & 42.46 & 42.46 & 65.30\,/\,71.02 & 61.06\,/\,58.99 & 0.811 & 0.244 \\
T+A & 45.53 & 51.31 & 85.16\,/\,83.19 & 84.97\,/\,83.07 & 0.739 & 0.790 & 52.29 & 53.68 & 85.77\,/\,84.65 & 85.73\,/\,84.20 & 0.549 & 0.764 \\
T+V & 45.40 & 51.31 & 84.86\,/\,82.60 & 84.70\,/\,82.56 & 0.737 & 0.790 & 52.48 & 53.96 & 86.08\,/\,84.85 & 85.99\,/\,84.85 & 0.545 & 0.768 \\
A+V & 22.40 & 24.30 & 59.25\,/\,58.31 & 56.45\,/\,57.04 & 1.350 & 0.198 & 42.52 & 42.52 & 65.38\,/\,71.22 & 60.40\,/\,60.12 & 0.812 & 0.244 \\
\midrule
\rowcolor{green!10}
Random & 34.52 & 38.55 & 73.27/71.93 & 73.03/71.93 & 1.046 & 0.532 
& 47.10 & 47.94 & 78.20/78.94 & 78.46/80.00 & 0.664 & 0.595 \\
\bottomrule
\end{tabular}
}
\end{table*}

\subsection{Effects of Different Modalities}
To assess the contribution of each modality under incomplete conditions, we conduct modality ablation experiments on MOSI and MOSEI. During training, random modality dropping is applied to simulate incomplete inputs, while at test time we evaluate three unimodal settings (T, A, and V), three bimodal settings (T+A, T+V, and A+V), and a random-missing setting (Random), with results reported in Table~\ref{tab:results3}. The results show that language is the most informative modality on both datasets, as text-involved settings consistently outperform A, V, and A+V, indicating that linguistic information provides the primary basis for sentiment prediction. In contrast, the relatively weak performance of non-textual settings suggests that audio and visual cues alone are insufficient for reliable sentiment understanding. The strong results of T+A and T+V further indicate that these modalities mainly serve as complementary signals when language is available. Under the random-missing setting, our model still maintains strong performance, demonstrating its robustness to stochastic modality absence.

\begin{figure}
    \centering
     \setlength{\abovecaptionskip}{2pt}
    \setlength{\belowcaptionskip}{-6pt}
    \includegraphics[width=1\linewidth]{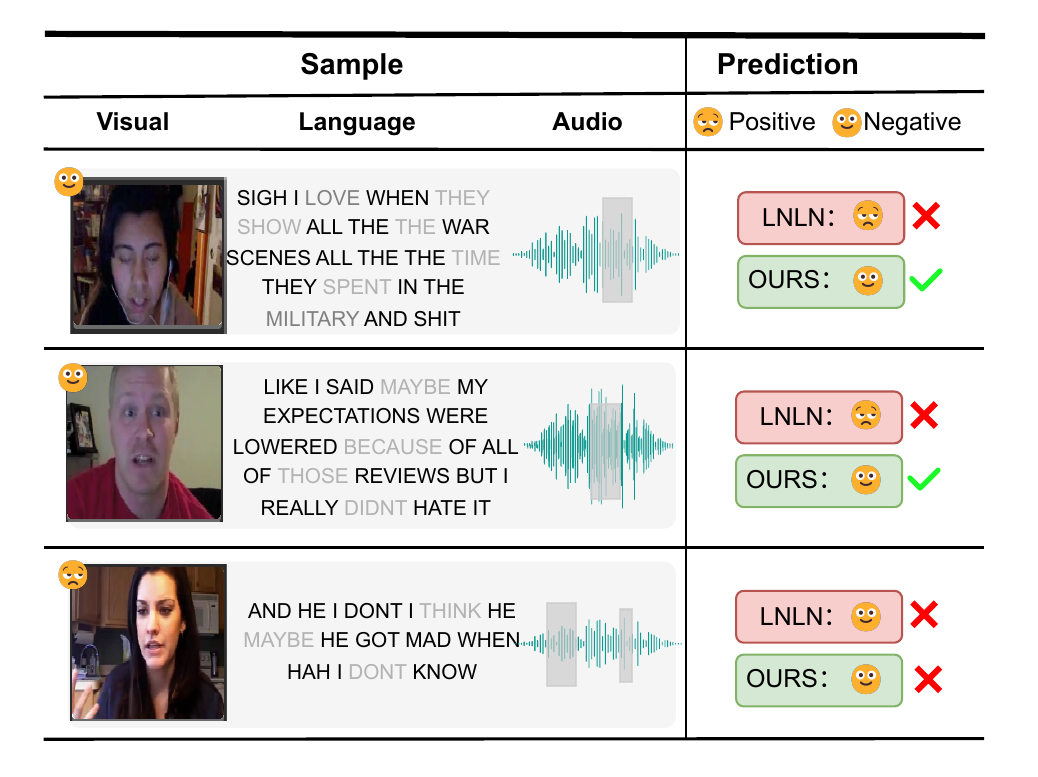}
    \caption{Case study on three MOSI samples. Our method correctly predicts the first two cases where LNLN fails, suggesting its advantage in handling incomplete multimodal cues. Both methods fail on the third case, revealing the difficulty of highly ambiguous samples.}
    \label{case}
\end{figure}

\subsection{Effects of Different Components}

To assess the contribution of each component, we conduct ablation studies on MOSI and SIMS. FULL denotes the complete model. w/o Proxy removes the proxy branch and directly predicts from the observed language representation. w/o Iterative disables iterative correction and uses only a one-shot proxy. w/o $\mathcal{L}_{corr}$ removes the stage-wise correction loss while keeping the rest of the framework unchanged. All variants are trained and evaluated under the same settings, and the results are reported in Table~\ref{tab:AblationComparison1}. Overall, FULL achieves the most balanced performance, especially on SIMS, where all three ablated variants—w/o Proxy, w/o Iterative, and w/o $\mathcal{L}_{corr}$—show clear and consistent degradation. Among them, w/o Proxy leads to the largest drop, highlighting the importance of proxy-based compensation under incomplete conditions. Both w/o Iterative and w/o $\mathcal{L}_{corr}$ also underperform FULL, indicating that iterative refinement and stage-wise correction supervision are both beneficial for improving proxy quality. On MOSI, the gains are relatively smaller, but FULL still achieves the best results on the main classification metrics and remains competitive on the others. These results confirm that each proposed component contributes to the effectiveness of the overall framework.

\begin{table*}[t]
\centering
\caption{Component ablation results on the MOSI and SIMS datasets}
\label{tab:AblationComparison1}
\setlength{\tabcolsep}{5.5pt}
\renewcommand{\arraystretch}{1.08}
\resizebox{\linewidth}{!}{%
\begin{tabular}{lcccccccccccc}
\toprule
\multirow{2}{*}{Method} 
& \multicolumn{6}{c}{\textbf{MOSI}} 
& \multicolumn{6}{c}{\textbf{SIMS}} \\
\cmidrule(lr){2-7}\cmidrule(lr){8-13}
& Acc-7  & Acc-5  & Acc-2  & F1  & MAE  & Corr 
& Acc-5  & Acc-3  & Acc-2  & F1  & MAE  & Corr  \\
\midrule
\rowcolor{green!10}
\textbf{FULL} 
& 34.52 & 38.55 & \textbf{73.27}/\textbf{71.93} & \textbf{73.03}/\textbf{71.93} & \textbf{1.046} & 0.532  
& \textbf{35.13} & \textbf{58.28} & \textbf{73.05} & \textbf{76.29} & \textbf{0.498} & \textbf{0.404} \\
\midrule
w/o Proxy 
& \textbf{34.53} & \textbf{38.57} & 72.29\,/\,71.89 & 72.29\,/\,71.84 & 1.053 & 0.527
& 30.93 & 54.98 & 70.05 & 62.08 & 0.569 & 0.238 \\
w/o Iterative
& 34.27 & 38.32 & 73.19\,/\,71.84 & 72.37\,/\,71.69 & 1.051 & \textbf{0.536}
& 30.98 & 55.23 & 70.58 & 69.74 & 0.565 & 0.241 \\
w/o $\mathcal{L}_{corr}$
& 34.23 & 38.25 & 72.37\,/\,71.29 & 71.48\,/\,70.45 & 1.051 & 0.529
& 31.05 & 55.03 & 70.09 & 62.12 & 0.570 & 0.236 \\
\bottomrule
\end{tabular}%
}
\end{table*}

\subsection{Case Study}
As shown in Figure~\ref{case}, we present three case studies from MOSI. In the first two cases, our method yields correct predictions whereas LNLN fails, showing that iterative proxy correction can better recover sentiment-relevant semantics from incomplete inputs by leveraging complementary multimodal cues. In the third case, both methods produce incorrect predictions, indicating that severe ambiguity or conflicting cross-modal evidence remains challenging. These examples qualitatively demonstrate the advantage of our method in handling incomplete multimodal inputs, while also revealing the difficulty of extremely hard cases.

\section{Conclusion}
In this paper, we investigate incomplete multimodal sentiment analysis from the perspective of proxy reliability. Existing proxy-based methods typically rely on one-shot proxy construction, making downstream reasoning sensitive to initial proxy errors. To address this limitation, we propose an iterative proxy correction framework that progressively refines a language-oriented proxy under multimodal context before prediction. The proxy is initialized from non-language modalities, updated through gated residual correction, and adaptively fused with the observed language representation via an estimated reliability score. We further introduce a stage-wise correction objective to regularize the refinement process. Extensive experiments verify the effectiveness and robustness of the proposed method under incomplete conditions.

\bibliographystyle{IEEEtran}
\bibliography{bib/ref}

@article{liu2026,
title = {Multimodal sentiment analysis based on label semantic guidance under social links},
journal = {Pattern Recognition},
volume = {171},
pages = {112277},
year = {2026},
doi = {https://doi.org/10.1016/j.patcog.2025.112277},
author = {Yun Liu and Xiaoming Zhang and Bo Zhang and others},
}

@article{Wang-2025-PublicBehavior,
title        = {Public Behavior and Emotion Correlation Mining Driven by Aspect From News Corpus},
author       = {Wang, Xinzhi and Chang, Yudong and Kou, Luyao and others},
year         = 2025,
journal      = {{IEEE} Transactions on Neural Networks and Learning Systems},
volume       = 36,
number       = 5,
pages        = {8632--8645}
}

@article{AFROZE2026,
title = {MulMoSenT: Multimodal sentiment analysis for a low-resource language using textual-visual cross-attention and fusion},
journal = {Information Fusion},
volume = {131},
pages = {104129},
year = {2026},
issn = {1566-2535},
doi = {https://doi.org/10.1016/j.inffus.2026.104129},
author = {Sadia Afroze and Md. Rajib Hossain and Mohammed Moshiul Hoque and others},

}

@article{ZHAI2026131746,
title = {Hybrid prompt learning and multilevel knowledge distillation for multimodal sentiment analysis with missing modalities},
journal = {Expert Systems with Applications},
volume = {315},
pages = {131746},
year = {2026},
issn = {0957-4174},
doi = {https://doi.org/10.1016/j.eswa.2026.131746},
author = {Yiqiao Zhai and Qiuxia Yang and Chengchao Wang and others},

}

@article{ZHAO2025126974,
title = {Toward Robust Multimodal Sentiment Analysis using multimodal foundational models},
journal = {Expert Systems with Applications},
volume = {276},
pages = {126974},
year = {2025},
issn = {0957-4174},
doi = {https://doi.org/10.1016/j.eswa.2025.126974},
author = {Xianbing Zhao and Soujanya Poria and Xuejiao Li and others},

}

@inproceedings{guo2024multimodal,
  title={Multimodal prompt learning with missing modalities for sentiment analysis and emotion recognition},
  author={Guo, Zirun and Jin, Tao and Zhao, Zhou},
  year={2024},
  booktitle = "Proceedings of the 62nd Annual Meeting of the Association for Computational Linguistics (Volume 1: Long Papers)",
  pages = "1726--1736",
  doi={https://doi.org/10.18653/v1/2024.acl-long.94},
}

@article{GONG2026111870,
title = {Towards robust sentiment analysis with multimodal interaction graph and hybrid contrastive learning},
journal = {Pattern Recognition},
volume = {169},
pages = {111870},
year = {2026},
issn = {0031-3203},
doi = {https://doi.org/10.1016/j.patcog.2025.111870},
author = {Peizhu Gong and Jin Liu and Xiliang Zhang and others},

}

@inproceedings{lin-etal-2025-t2dr,
    title = "$T^2DR$: A Two-Tier Deficiency-Resistant Framework for Incomplete Multimodal Learning",
    author = "Lin, Han  and
      Tang, Xiu  and
      Li, Huan  and
      others",
    booktitle = "Findings of the Association for Computational Linguistics: ACL 2025",
    month = jul,
    year = "2025",
    address = "Vienna, Austria",
    publisher = "Association for Computational Linguistics",
    doi = "10.18653/v1/2025.findings-acl.452",
    pages = "8602--8616",
}

@article{wang2025sslmm,
  title={SSLMM: Semi-Supervised Learning with Missing Modalities for Multimodal Sentiment Analysis},
  author={Wang, Yiyu and Jian, Haifang and Zhuang, Jian and others},
  year={2025},
  journal={Information Fusion},
  volume={120},
  pages={103058},
  publisher={Elsevier},
doi={https://doi.org/10.1016/j.inffus.2025.103058},
}

@inproceedings{li2024unified,
  title={A unified self-distillation framework for multimodal sentiment analysis with uncertain missing modalities},
  author={Li, Mingcheng and Yang, Dingkang and Lei, Yuxuan and others},
  year={2024},
  booktitle={Proceedings of the AAAI conference on artificial intelligence},
  pages={10074--10082},
  doi={https://doi.org/10.1609/aaai.v38i9.28871},
}

@inproceedings{hazarika2020misa,
  title={Misa: Modality-invariant and-specific representations for multimodal sentiment analysis},
  author={Hazarika, Devamanyu and Zimmermann, Roger and Poria, Soujanya},
  year={2020},
  booktitle={Proceedings of the 28th ACM international conference on multimedia},
  pages={1122--1131},
  doi={https://doi.org/10.1145/3394171.3413678},
}

@inproceedings{zhu-etal-2025-proxy,
    title = "Proxy-Driven Robust Multimodal Sentiment Analysis with Incomplete Data",
    author = "Zhu, Aoqiang  and
      Hu, Min  and
      Wang, Xiaohua  and
      others",
    booktitle = "Proceedings of the 63rd Annual Meeting of the Association for Computational Linguistics (Volume 1: Long Papers)",
    month = jul,
    year = "2025",
    publisher = "Association for Computational Linguistics",
    doi = "10.18653/v1/2025.acl-long.1075",
    pages = "22123--22138",
}

@inproceedings{zhang2024towards,
title={Towards Robust Multimodal Sentiment Analysis with Incomplete Data},
author={Haoyu Zhang and Wenbin Wang and Tianshu Yu},
year={2024},
booktitle={The Thirty-eighth Annual Conference on Neural Information Processing Systems},
pages = {55943--55974},
doi={https://doi.org/10.48550/arXiv.2409.20012},
}

@inproceedings{Luan2017cascaded,
  title={Missing Modalities Imputation via Cascaded Residual Autoencoder}, 
  author={Tran, Luan and Liu, Xiaoming and Zhou, Jiayu and others},
  booktitle={2017 IEEE Conference on Computer Vision and Pattern Recognition (CVPR)}, 
  year={2017},
  pages={4971-4980},
  doi={https://doi.org/10.1109/CVPR.2017.528}}

@inproceedings{shang2017vigan,
  title={VIGAN: Missing view imputation with generative adversarial networks},
  author={Shang, Chao and Palmer, Aaron and Sun, Jiangwen and others},
  year={2017},
  booktitle={2017 IEEE International conference on big data (Big Data)},
  pages={766--775},
doi={https://doi.org/ 10.1109/BigData.2017.8257992}
}

@article{zeng2022robust,
  title={Robust multimodal sentiment analysis via tag encoding of uncertain missing modalities},
  author={Zeng, Jiandian and Zhou, Jiantao and Liu, Tianyi},
  year={2022},
  journal={IEEE Transactions on Multimedia},
  volume={25},
  pages={6301--6314},
  doi={https://doi.org/10.1109/TMM.2022.3207572}
}

@inproceedings{yu2021learning,
  title={Learning modality-specific representations with self-supervised multi-task learning for multimodal sentiment analysis},
  author={Yu, Wenmeng and Xu, Hua and Yuan, Ziqi and others},
  year={2021},
  booktitle={Proceedings of the AAAI conference on artificial intelligence},
  pages={10790--10797},
  doi={https://doi.org/10.1609/aaai.v35i12.17289},
}

@article{wang2022cross,
  title={Cross-modal enhancement network for multimodal sentiment analysis},
  author={Wang, Di and Liu, Shuai and Wang, Quan and others},
  year={2022},
  journal={IEEE Transactions on Multimedia},
  volume={25},
  pages={4909--4921},
  doi={https://doi.org/10.1109/TMM.2022.3183830},
}

@article{wang2023tetfn,
  title={TETFN: A text enhanced transformer fusion network for multimodal sentiment analysis},
  author={Wang, Di and Guo, Xutong and Tian, Yumin and others},
  year={2023},
  journal={Pattern Recognition},
  volume={136},
  pages={109259},
  doi={https://doi.org/10.1016/j.patcog.2022.109259},
}

@inproceedings{han2021improving,
  title={Improving multimodal fusion with hierarchical mutual information maximization for multimodal sentiment analysis},
  author={Han, Wei and Chen, Hui and Poria, Soujanya},
  year = {2021},
  booktitle = "Proceedings of the 2021 Conference on Empirical Methods in Natural Language Processing",
  pages = {9180--9192},
  doi={https://doi.org/10.18653/v1/2021.emnlp-main.723},
}

@inproceedings{zhang2023learning,
  title={Learning language-guided adaptive hyper-modality representation for multimodal sentiment analysis},
  author={Zhang, Haoyu and Wang, Yu and Yin, Guanghao and others},
  year = {2023},
  booktitle = "Proceedings of the 2023 Conference on Empirical Methods in Natural Language Processing",
  pages = {756--767},
  doi={https://doi.org/10.18653/v1/2023.emnlp-main.49},
}

@article{zheng20263d,
  title={3d smoke scene reconstruction guided by vision priors from multimodal large language models},
  author={Zheng, Xinye and Wang, Fei and Nie, Yiqi and others},
  journal={arXiv preprint arXiv:2604.05687},
  year={2026}
}

@article{li2025tactic,
  title={Tactic: Translation agents with cognitive-theoretic interactive collaboration},
  author={Li, Weiya and Chen, Junjie and Li, Bei and others},
  journal={arXiv preprint arXiv:2506.08403},
  year={2025}
}

\end{document}